\documentclass[conference]{IEEEtran}

\usepackage{cite}
\usepackage{amsmath,amssymb,amsfonts}
\usepackage{algorithmic}
\usepackage{graphicx}
\usepackage{textcomp}
\usepackage{xcolor}
\usepackage{multirow}
\usepackage[hidelinks]{hyperref}
\usepackage{tabularx}
\usepackage{subcaption}
\usepackage{pgfplots}
\pgfplotsset{compat=1.17}

\begin{document}

\IEEEoverridecommandlockouts
\IEEEpubid{\makebox[\columnwidth]{ 979-8-3503-7608-1/24\$31.00 \copyright2024 IEEE \hfill} \hspace{\columnsep}\makebox[\columnwidth]{}}

\title{RTL-Repo: A Benchmark for Evaluating LLMs on Large-Scale RTL Design Projects}

\author{\IEEEauthorblockN{Ahmed Allam}
\IEEEauthorblockA{\textit{Department of Computer Science and Engineering} \\
\textit{The American University in Cairo}\\
Cairo, Egypt \\
ahmedeallam@aucegypt.edu}

\and

\IEEEauthorblockN{Mohamed Shalan}
\IEEEauthorblockA{\textit{Department of Computer Science and Engineering} \\
\textit{The American University in Cairo}\\
Cairo, Egypt \\
mshalan@aucegypt.edu
}}

\IEEEaftertitletext{\vspace{-2.1\baselineskip}}

\maketitle

\begin{abstract}

Large Language Models (LLMs) have demonstrated potential in assisting with Register Transfer Level (RTL) design tasks. Nevertheless, there remains to be a significant gap in benchmarks that accurately reflect the complexity of real-world RTL projects. To address this, this paper presents RTL-Repo, a benchmark specifically designed to evaluate LLMs on large-scale RTL design projects. RTL-Repo includes a comprehensive dataset of more than 4000 Verilog code samples extracted from public GitHub repositories, with each sample providing the full context of the corresponding repository. We evaluate several state-of-the-art models on the RTL-Repo benchmark, including GPT-4, GPT-3.5, Starcoder2, alongside Verilog-specific models like VeriGen and RTLCoder, and compare their performance in generating Verilog code for complex projects. The RTL-Repo benchmark provides a valuable resource for the hardware design community to assess and compare LLMs' performance in real-world RTL design scenarios and train LLMs specifically for Verilog code generation in complex, multi-file RTL projects. RTL-Repo is open-source and publicly available on Github\footnote{\url{https://github.com/AUCOHL/RTL-Repo}}.

\end{abstract}


\section{Introduction}

In a typical ASIC flow, RTL modeling and verification are labor-intensive manual processes, contrasting with other design steps that are fully automated using Electronic Design Automation (EDA) tools. Consequently, these manual tasks are prone to consuming significant time and are susceptible to errors. Traditionally, the hardware design community has sought to address these challenges by raising the abstraction level by adopting high-level languages like C to describe hardware behavior instead of its microarchitecture \cite{Dynamatic}. However, this approach entails reliance on high-level synthesis tools, potentially compromising hardware efficiency. A recent strategy to relieve this issue involves leveraging artificial intelligence (AI) in the form of Large Language Models (LLMs) to expedite Register Transfer Level (RTL) modeling and verification, thereby reducing time-to-market. LLMs have already demonstrated their effectiveness in generating high-quality code for programming languages such as Python, Java, and C++ \cite{opencodeinterpreter,nijkamp2023codegen,lozhkov2024starcoder}. On the other hand, the application of LLMs to hardware descriptive languages like Verilog remains under-explored, presenting a significant opportunity for innovation in the hardware design domain.

Recent research has focused on fine-tuning existing models to generate Verilog code. An example of such models is VeriGen \cite{thakur2023verigen}, a 16B parameter model based on CodeGen, fine-tuned on a large dataset of GitHub code and Verilog textbooks.  Another example is RTLCoder \cite{liu2024rtlcoder}, a 7B parameter model built upon Mistral and DeepSeek, showcasing better performance over GPT-3.5 while being lightweight. In addition, \cite{liu2024chipnemo} introduced ChipNeMo, a model based on LLaMA that excels in engineering assistant chatbot and EDA script generation tasks. These models have shown promising results in generating Verilog code from high-level specifications, making them valuable tools for hardware design automation.

As the field of LLMs for hardware design continues to evolve, there is a need for benchmarks and datasets capable of evaluating the performance of these models in generating Verilog code. Several benchmarks have been proposed recently, such as RTLLM \cite{lu2023rtllm} and VerilogEval \cite{liu2023verilogeval}. However, these benchmarks prompt the model to generate a single standalone Verilog module, which does not reflect the real-world application of generating Verilog RTL models in large and complex projects and does not measure the model's ability to understand and generate code in multi-file contexts. This highlights the pressing need for a new benchmark to evaluate model performance in more realistic RTL project environments.

In this paper, we present RTL-Repo, an open-source benchmark designed to evaluate the performance of Large Language Models (LLMs) in generating Verilog code in multi-file, large-scale projects in real-world RTL design scenarios. The benchmark consists of a dataset of over 4000 code samples extracted from public GitHub repositories, each containing the context of all Verilog files. Compared to existing benchmarks, RTL-Repo provides a more realistic and challenging evaluation of the model's performance in RTL design, in addition to being orders of magnitude larger in terms of dataset size and context length. We conduct a comprehensive evaluation of several state-of-the-art models on the RTL-Repo benchmark, including GPT-4, GPT-3.5, Starcoder2, VeriGen, and RTLCoder, and compare their performance in generating Verilog code in large-scale projects. Additionally, we provide a training dataset that can be used to fine-tune new models to better handle long-range dependencies and multi-file contexts in RTL code generation.

\section{Related Work}

Recent benchmarks for evaluating LLMs in real-world complex software engineering tasks include notable examples such as CrossCodeEval \cite{ding2023crosscodeeval} and RepoBench \cite{liu2023repobench}. CrossCodeEval challenges the model to generate code snippets in Python, Java, TypeScript, and C\# by requiring an in-depth cross-file contextual understanding. RepoBench evaluates models for repository-level code completion in Python and Java, measuring the ability to predict the next line of code given the repository context. While these benchmarks represent significant progress, there is still a lack of benchmarks for evaluating LLMs in hardware description languages like Verilog, particularly for multi-file, large-scale codebases. This gap is crucial for real-world RTL design scenarios, where handling complex hardware design tasks is essential.

For evaluating LLMs in generating Verilog code, several benchmarks have been proposed \cite{lu2023rtllm, liu2023verilogeval, thakur2022benchmarking}. RTLLM \cite{lu2023rtllm} is benchmarked with 29 RTL design tasks described in natural language, and evaluated based on the syntactic and functional correctness of the generated Verilog code. VerilogEval \cite{liu2023verilogeval} contains 156 problems extracted from HDLBits, focusing on the correctness of the generated code. These benchmarks assess the models' performance in single-file Verilog code generation, which does not accurately represent the real-world scenario of generating Verilog code across large codebases. Additionally, the limited number of samples in these benchmarks restricts task diversity, making it easier for models to memorize specific solutions rather than demonstrating true generalizability and robustness in varied and complex real-world applications. This highlights the need for more comprehensive benchmarks that evaluate LLMs in the context of large and complex Verilog codebases, addressing real-world requirements and challenges.

\section{RTL-Repo}

RTL-Repo is a benchmark for evaluating the effectiveness of LLMs in generating Verilog code autocompletions in large and complex codebases. The benchmark aims to evaluate the model's ability to remember and understand the context of the entire Verilog repository and incorporate this knowledge into generating new code that is correct, relevant, consistent in terms of logic as well as coding conventions and guidelines, and aware of all other components and modules in the project. Furthermore, it evaluates how well models handle long context inputs, addressing a common limitation where many LLMs tend to struggle to utilize context information, hallucinate, and degrade in performance after a certain number of tokens \cite{li2024longcontext, an2024make}. This way, we test the needed abilities that any LLM must have in order to be useful in real-world settings where work is done on a large project, not just a short, standalone module. This provides a more accurate, realistic quantitative evaluation of a model's performance in real-world RTL design scenarios.

\subsection{Benchmark Construction}

To construct the RTL-Repo dataset, we use the Github API to collect all public Verilog repositories. Repositories were collected only if they had a permissive license and were created after October 15, 2023, to prevent data leakage with newer training datasets like Stack V2 used by models such as Starcoder2 \cite{lozhkov2024starcoder}. We filter the repositories to include those with at least four Verilog files and a maximum of 24 files. This ensures that the repositories are large enough to provide a comprehensive context for the model to learn from while also being manageable to avoid being too complex for the model to handle.

After collecting the repositories, we extract all the Verilog files from each one. Then we choose four random files from each repository, and for each file, we choose a random, non-empty, non-comment line of code to be the target line for the model to predict. Each sample in the dataset contains the context of the entire repository, the code of the current file being edited, excluding the target line and all lines after it, and the target line itself. After constructing the dataset, we split it into training and test splits. The training split of the dataset can be used later to fine-tune new models to be better at handling long-range dependencies and multi-file contexts in Verilog code generation.

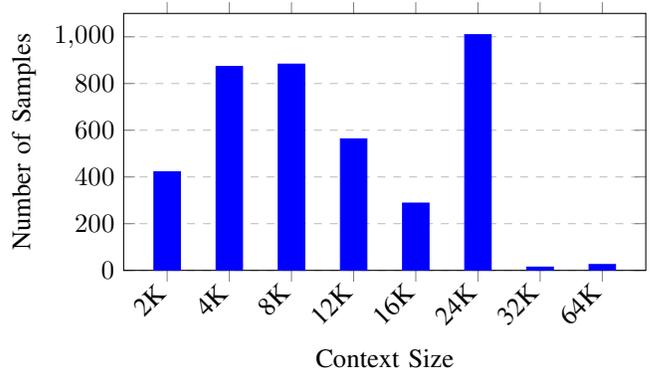
\begin{figure}[!t]
  \centering
  \begin{tikzpicture}
    \begin{axis}[
        ybar,
        xlabel={Context Size},
        ylabel={Number of Samples},
        symbolic x coords={2K,4K,8K,12K,16K,24K,32K,64K},
        xtick=data,
        x tick label style={rotate=45, anchor=east},
        ymin=0,
        ymax=1100,
        ytick={0, 200, 400, 600, 800, 1000},
        legend pos=north west,
        ymajorgrids=true,
        grid style=dashed,
        width=0.47\textwidth,
        height=5cm,
    ]

    \addplot[
        fill=blue!30,
        color=blue,
        mark=dot,
        ]
        coordinates {
        (2K, 422)
        (4K, 873)
        (8K, 883)
        (12K, 563)
        (16K, 288)
        (24K, 1009)
        (32K, 14)
        (64K, 25)
        };

    \end{axis}
  \end{tikzpicture}
  \caption{Distribution of context sizes in the RTL-Repo Benchmark Dataset, illustrating the frequency of different context lengths in the dataset samples.}
  \label{fig:RTLRepo_stats}
  \vspace{-10pt}
\end{figure}

The dataset contains 4098 samples extracted from 1361 public GitHub repositories. The training split contains 2924 samples, while the test split contains 1174 samples. We use the GPT-4 tokenizer to measure the number of tokens in the context of each sample, and we split the dataset into nine categories based on it, ranging from 2K to 128K tokens. The average number of tokens in the context of the samples is 12.6K tokens. The distribution of context sizes in the dataset is shown in Figure \ref{fig:RTLRepo_stats}.

\subsection{Task Formulation}

In the task of the RTL-Repo benchmark, a model is given the context of the entire repository and the lines before the target line in the current file. Similar to \cite{liu2023repobench}, the model is tasked with predicting the target line of code. The task can be formally defined as follows:

\begin{equation}
    P(Y) = \prod_{i=1}^{n} P(y_i | y_{<i}, \mathit{C}_{repo}, \mathit{C}_{file})
\end{equation}

where $P(Y)$ is the joint probability of the predicted sequence $Y$. Each token $y_i$ in $Y$ is predicted based on its preceding tokens $y_{<i}$, the context of the entire repository $\mathit{C}_{repo}$, and the context of the current file edited $\mathit{C}_{file}$, reflecting the autoregressive nature of the task.

Evaluating model performance by predicting a single line is effective because it directly assesses the model's capability to manage specific, localized code generation tasks within the broader context of the entire repository, while also being straightforward to measure. This approach effectively tests the model's understanding of the code structure and logic as well as its capability to retain and utilize relevant information from a large context to generate new code. This evaluation task ensures practical relevance for a hardware design coding assistant, focusing on the model’s proficiency in providing accurate and contextually correct code completions.

\subsection{Evaluation Metrics}

We use both Exact Match (EM) and Edit Similarity (ES) as evaluation metrics for the RTL-Repo benchmark. The EM metric measures the percentage of samples where the model's prediction exactly matches the target line of code, while the ES metric measures the average edit similarity between the model's prediction and the target line of code, calculated using the Levenshtein distance between the two lines of code. Using similarity to evaluate a model's performance is sufficient for single-line code completion tasks since it effectively measures the effort required by developers to correct errors within an autocompletion system \cite{codexglue}. These metrics directly reflect the practical utility and accuracy of the model in real-world coding scenarios.


\subsection{Comparison to Existing RTL Benchmarks}

Existing benchmarks for evaluating LLMs in generating Verilog code, such as RTLLM \cite{lu2023rtllm} and VerilogEval \cite{liu2023verilogeval}, focus on evaluating the model's performance in generating single Verilog files that are often small and do not interact with other components. In contrast, the RTL-Repo benchmark evaluates the model's performance in generating Verilog code in multi-file, large-scale codebases, providing a more realistic and challenging evaluation of the model's performance in real-world RTL design scenarios. We discuss the comparison of the benchmarks in more detail below:

\begin{itemize}

\item \textbf{Real-world RTL design scenarios:} Our benchmark evaluates the model's performance in RTL design scenarios that require a broad context and a deep understanding of real-world projects with large codebases, providing a more accurate evaluation of the model's performance in real-world applications. All projects in the RTL-Repo dataset are real-world projects from public GitHub repositories, ensuring that the model is evaluated on realistic and challenging tasks.

\item \textbf{Multi-file codebase input:} In each task, the model is provided with the context of the entire repository, which challenges the model to remember and understand the context of multiple files within a project and incorporate this knowledge into the task of generating new code in any file. This provides a more comprehensive evaluation of the model's performance in generating Verilog code in large-scale codebases, unlike existing benchmarks focusing on generating only a single module.

\item \textbf{Large diverse RTL tasks:} We compare the RTL-Repo benchmark with existing RTL benchmarks regarding dataset size and average context length in Table \ref{tbl:benchmarks}. We calculate the context length for the model's input and expected output using the GPT-4 tokenizer. The RTL-Repo benchmark contains 1174 samples with an average context length of 12.6K tokens, providing a large and diverse dataset for evaluating the model's performance in generating Verilog code in large-scale codebases. In contrast, VerilogEval-Human contains 156 samples with an average context length of 266 tokens, VerilogEval-Machine contains 143 samples with an average context length of 292 tokens, and RTLLM contains 29 samples with an average context length of 895 tokens, which limits the diversity of tasks and scenarios on which the model is evaluated.

\item \textbf{Living benchmark:} As the process of collecting the dataset is fully automated, the benchmark can be easily extended to include more samples from new repositories, ensuring that the benchmark remains up-to-date and not have any data leakage with any dataset that a new model may be trained on in the future. We plan to keep our benchmark a living one by updating it regularly.

\end{itemize}

\begin{table}[h!]
  \centering
  \renewcommand{\arraystretch}{1.4}
  \begin{tabular}{|c|c|c|}
    \hline
    Benchmark & Dataset Size & Avg. Context Length \\
    \hline
    VerilogEval-Human \cite{liu2023verilogeval} & 156 & 266 \\
    VerilogEval-Machine \cite{liu2023verilogeval} & 143 & 292 \\
    RTLLM \cite{lu2023rtllm} & 29 & 895 \\
    RTL-Repo & \textbf{1174} & \textbf{12.6K} \\
    \hline
  \end{tabular}
  \caption{Comparison of Benchmark Datasets in terms of number of samples and average aontext length}
  \label{tbl:benchmarks}
\end{table}

\section{Experimental Design}

\subsection{Models Evaluated}

In our experiments, we evaluate several state-of-the-art models on the RTL-Repo benchmark. We assess closed-source models, such as GPT-3.5 and GPT-4, through the OpenAI API. We also evaluate open-source models, including Starcoder2 \cite{lozhkov2024starcoder}, which contains 15B parameters, and models specifically trained for Verilog code generation, such as RTLCoder-Mistral and RTLCoder-DeepSeek, which are 7B parameter models \cite{liu2024rtlcoder}, and VeriGen \cite{thakur2023verigen}, a 16B parameter model based on Codegen \cite{nijkamp2023codegen}, trained on a large dataset of Verilog GitHub code and textbooks.

\newcolumntype{L}[1]{>{\centering\let\newline\\\arraybackslash\hspace{0pt}}m{#1}}

\begin{table*}[!t]
  \centering
  \renewcommand{\arraystretch}{1.5}
  \small
  \begin{tabular}{|c||c|c|L{2cm}|L{2cm}|}
    \hline
   \multirow{2}{*}{Model} & \multirow{2}{*}{Type} & \multirow{2}{*}{\shortstack{Num of \\ Parameters}} & \multirow{2}{*}{Edit Similarity} & \multirow{2}{*}{Exact Match} \\
    & & & & \\
    \hline
    GPT-3.5 & \multirow{3}{*}{General-Purpose} & N/A & 61.4 & 33.8 \\
    GPT-4 \footnotemark & & N/A & \textbf{71.87} & \textbf{48.5} \\
    Starcoder2 \cite{lozhkov2024starcoder} & & 15B & 48.0 & 17.0 \\
    \hline
    VeriGen \cite{thakur2023verigen} & \multirow{3}{*}{Verilog-Specific} & 16B & 43.9 & 9.5 \\
    RTLCoder-Mistral \cite{liu2024rtlcoder} & & 7B & 37.9 & 8.3 \\
    RTLCoder-DeepSeek \cite{liu2024rtlcoder} & & 6.7B & 48.1 & 16.2 \\
    \hline
  \end{tabular}
  \vspace{2pt}
  \caption{Performance Comparison of Various Models on the RTL-Repo Benchmark.}
  \label{tbl:results}
  \vspace{-10pt}
\end{table*}

\subsection{Approach}

For the evaluations, each model was prompted to generate a random target line of code, given the entire repository’s context and the lines preceding the target line in the current file. For generation, we used sampling with a temperature of 0.2, a top-k value of 50, and a top-p value of 1.0. We used a maximum context size for inputs consistent with the maximum for each model: 128K for GPT-4, 16K for Starcoder2 and GPT-3.5, and 2048 for RTLCoder and VeriGen. Models with smaller maximum context sizes are at a disadvantage for samples that exceed their context limit, as they are less capable of retaining and processing the entire repository's context. However, this reflects the practical limitations these models face in real-world applications, where handling large contexts is crucial for effective code generation.

\section{Results}
\label{sec:results}

\begin{figure}[!t]
  \centering
  \begin{tikzpicture}
    \begin{axis}[
        ybar,
        xlabel={Context Size},
        ylabel={Edit Similarity},
        symbolic x coords={2K,4K,8K,16K,24K,32K,64K},
        xtick=data,
        x tick label style={rotate=45, anchor=east},
        x=1.05cm,
        ymin=20,
        ymax=75,
        ytick={0, 10, 20, 30, 40, 50, 60, 70},
        legend style={at={(0.5,1.3)},
        anchor=north,legend columns=-1},
        ymajorgrids=true,
        grid style=dashed,
        width=0.5\textwidth,
        height=5cm,
        bar width=6pt,
        legend image code/.code={
            \draw [#1] (0cm,-0.1cm) rectangle (0.2cm,0.2cm);
        }
        ]

        \addplot[
            fill=green!30,
            color=green,
            mark=dot,
            ]
            coordinates {
            (2K, 56.1)
            (4K,  54.3)

            (8K, 51.3)

            (16K, 46.2)
            (24K, 38.7)

            (32K, 32.8)
            (64K, 28)

            };

        \addlegendentry{Starcoder2 \;\;}

        \addplot[
            fill=orange!30,
            color=orange,
            mark=dot,
            ]
            coordinates {
            (2K, 56.5)
            (4K,  54.5)

            (8K, 45.8)

            (16K, 48.6)
            (24K, 40.5)

            (32K, 39.5)
            (64K, 26.1)

            };

        \addlegendentry{RTLCoder-Deepseek \;\;}

        \addplot[
            fill=blue!30,
            color=blue,
            mark=dot,
            ]
            coordinates {
            (2K, 70.9)
            (4K,  68.5)

            (8K, 65.0)

            (16K, 64.4)
            (24K, 48.2)

            (32K, 44.3)
            (64K, 32.6)

            };

        \addlegendentry{GPT-3.5}
    \end{axis}
  \end{tikzpicture}
  \caption{Edit Similarity of GPT3.5, Starcoder2, and RTLCoder-Deepseek on the RTL-Repo benchmark for different context sizes}
  \label{fig:edit_similarity}
  \vspace{-10pt}
\end{figure}
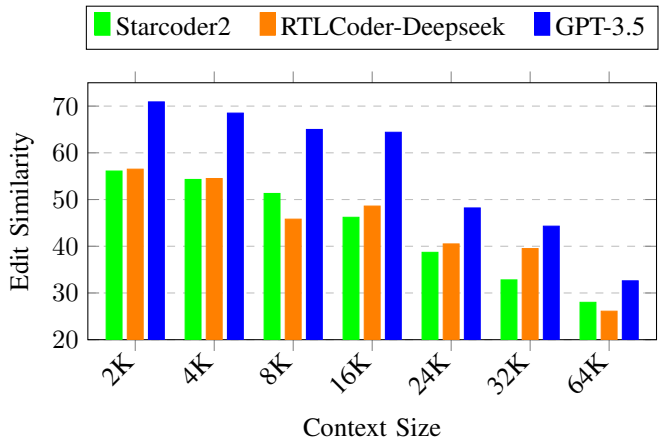

\footnotetext{Owing to budget limits, GPT-4 was evaluated on a smaller, random subset (200 samples out of 1174) of the dataset.}

The results of the models on the RTL-Repo benchmark are shown in Table \ref{tbl:results}. Results show that GPT-4 significantly outperforms all other models regarding both Edit Similarity and Exact Match of the Verilog code generated. GPT-4 achieves an Edit Similarity of 71.87 and an Exact Match of 48.5, while GPT-3.5 achieves an Edit Similarity of 61.4 and an Exact Match of 33.8. Starcoder2 is the best-performing open-source model in terms of Exact Match, achieving 17.0, while RTLCoder-DeepSeek is the best-performing open-source model in terms of Edit Similarity, achieving 48.1. VeriGen and RTLCoder-Mistral achieve the lowest performance in both metrics, with VeriGen achieving an Edit Similarity of 43.9 and an Exact Match of 9.5, and RTLCoder-Mistral achieving an Edit Similarity of 37.9 and an Exact Match of 8.3.

We analyzed the performance of the models across different context sizes to measure the correlation between context size and task difficulty. The results of the comparison between GPT3.5, Starcoder2, and RTLCoder-Deepseek are shown in Figure \ref{fig:edit_similarity}. We observe a significant decline in performance for all models as the context size increases. For samples in the 2K token range, GPT-3.5, Starcoder2, and RTL-Coder-DeepSeek achieve Edit Similarity scores of 70.9, 56.1, and 56.5, respectively. However, when the context size increases to 64K tokens, their performance drops considerably, with scores decreasing to 32.6, 28.0, and 26.1, respectively. This indicates that the models face greater difficulty handling long-range dependencies and multi-file contexts in Verilog. As the context size increases, it becomes more challenging for the models to retain and comprehend the entire repository's context and effectively incorporate this understanding into generating new code.

\section{Conclusion}

We presented RTL-Repo, a novel benchmark and dataset consisting of over 4000 code samples from  GitHub repositories, designed to evaluate the performance of Large Language Models (LLMs) in generating Verilog code within multi-file, large-scale RTL codebases. Our evaluation of state-of-the-art models, including GPT-4, GPT-3.5, Starcoder2, VeriGen, and RTLCoder, revealed that GPT-4 significantly outperformed all other models. In contrast, open-source Verilog-specific models struggled with long-range dependencies and multi-file contexts. The RTL-Repo benchmark not only provides a valuable resource for the hardware design community to assess and compare LLMs' performance in real-world RTL design scenarios but also serves as a foundation for future research in fine-tuning open-source models and developing more advanced LLMs that can effectively handle large RTL codebases.

\IEEEtriggeratref{8}
\bibliographystyle{IEEEtran}
\bibliography{main.bib}

\end{document}